# Multi-view Information-theoretic Co-clustering for Co-occurrence Data

Peng Xu[†], Zhaohong Deng[†*], Kup-Sze Choi[‡], Longbing Cao[§], Shitong Wang[†]

[†]School of Digital Media, Jiangnan University, China
[‡]Center of Smart Health, Hong Kong Polytechnic University, Hong Kong
[§]Advanced Analytics Institute, University of Technology Sydney, Australia
pengxujnu@163.com, dzh666828@aliyun.com, thomasks.choi@polyu.edu.hk, longbing.cao@uts.edu.au, wxwangst@aliyun.com

**Abstract**

Multi-view clustering has received much attention recently. Most of the existing multi-view clustering methods only focus on one-sided clustering. As the co-occurring data elements involve the counts of sample-feature co-occurrences, it is more efficient to conduct two-sided clustering along the samples and features simultaneously. To take advantage of two-sided clustering for the co-occurrences in the scene of multi-view clustering, a two-sided multi-view clustering method is proposed, i.e., multi-view information-theoretic co-clustering (MV-ITCC). The proposed method realizes two-sided clustering for co-occurring multi-view data under the formulation of information theory. More specifically, it exploits the agreement and disagreement among views by sharing a common clustering results along the sample dimension and keeping the clustering results of each view specific along the feature dimension. In addition, the mechanism of maximum entropy is also adopted to control the importance of different views, which can give a right balance in leveraging the agreement and disagreement. Extensive experiments are conducted on text and image multi-view datasets. The results clearly demonstrate the superiority of the proposed method.

## Introduction

With advances and development in multimedia, tremendous amount of unlabeled data with multiple attribute sets are generated on the Internet every day. They are called multi-view or multi-modal datasets (Ngiam et al. 2011). They are very common in real life, e.g., a webpage containing hyperlink-text and page-text, a document represented by multiple languages, or an image extracted with different feature descriptors (Li, Yang, and Zhang 2016).

The multi-view clustering is the most important method to deal with the unlabeled multi-view data. The foremost problem in multi-view clustering is how to exploit the agreement and disagreement among views simultaneously (Li, Yang, and Zhang 2016). Here the agreement and disagreement refer to consistency and complementary information among views respectively. It is crucial to get a right balance in leveraging the agreement and disagreement. Excessive excavation of the agreement would loss the complementary information among views and overemphasis on the disagreement would reduce the benefits from the common information extracted from the views. They can both degrade the algorithms (Xu, Tao, and Xu 2013). Most of existing algorithms tackled the problems from the aspect of collaborative strategies of multiple views, such as co-regularization (Kumar and Rai 2011), co-training (Kumar and Iii 2011) and multi-kernel learning (Guo et al. 2014), while neglected the characteristics of the datasets. Therefore, the existing multi-view clustering algorithms do not take full advantage of the characteristics of certain multi-view datasets, such as the sample-feature co-occurrence.

Co-occurrences widely exist in categorical data (Wang et al. 2011), natural language processing (Hotho et al. 2005), computer vision (Tsai 2012), and information retrieval (Rasiwasia et al. 2010), such as the text represented with bag of words model, the images represented with feature histograms, etc. The elements of co-occurring matrices (rows for samples and columns for features) can be viewed as the frequency or counts of a certain feature occurred in a certain sample. A basic problem for the co-occurrence analysis is co-clustering (two-sided clustering): simultaneous clustering of the samples and features. For the co-occurrences, two-sided clustering can perform more effectively than one-sided clustering (Dhillon and Guan 2003). Thus, in addition to the collaborated strategies of multiple views, two-sided clustering should be considered as well for the co-occurring data with multiple views.

To meet the above challenges, a novel method multi-view information-theoretic co-clustering (MV-ITCC) is proposed in the paper. The proposed method is based on information-theoretic co-clustering (ITCC), which is a two-sided clustering algorithm and it formalizes the clus-tering process as an optimization problem under the formulation of information theory (Dhillon et al. 2003).

---

[*]Corresponding author

More specifically, MV-ITCC simultaneously tackles the problems in multi-view clustering and achieves two-sided clustering from three aspects. First, it exploits the disagreement of different views by clustering the data of different views individually along the feature dimension. Next, the agreement of different views is exploited by clustering the concatenated data of all views along the sample dimension. Finally, MV-ITCC balances the agreement and disagreement by introducing maximum entropy into the objective function to control the importance of different views. Based on the mechanism of maximum entropy, each view is assigned with a weight during the optimization of the objective function adaptively. Different values of regularization parameter of the maximum entropy term can control the agreement and disagreement among views. The main contributions of the paper are summarized as follows:

- A two-sided multi-view clustering algorithm MV-ITCC is proposed. It can simultaneously exploit the agreement and disagreement for multi-view clustering and leverage the characteristics of the co-occurring data.
- The maximum entropy strategy is introduced into MV-ITCC to balance the agreement and disagreement among the views of multi-view datasets, which is crucial to ensure the success of multi-view clustering.
- An iterative algorithm is developed to solve the objective function of MV-ITCC and each iteration consists of three steps which correspond to the three aspects aforementioned.

Extensive experiments show that MV-ITCC significantly outperforms five popular multi-view clustering algorithms on seven datasets. And it shows obvious superiority on the high-dimensional and sparse textual data.

The remainder is organized as follows. The related work is briefly described in Section II. In Section III, the fundamentals of ITCC are first reviewed, then the proposed MV-ITCC and the optimization method are discussed in detail. Experimental studies on various datasets are reported in Section IV. Finally, the conclusions are given in Section V.

## Related Work

Most clustering algorithms focus on one-sided clustering, i.e., clustering one dimension of the data matrix based on similarities along another dimension. ITCC (Dhillon et al. 2003) clusters two dimensions simultaneously according to the loss of the mutual information between two variables corresponding to two dimensions of the data matrix respectively. This idea is consistent with information bottleneck theory (Tishby, Pereira, and Bialek. 2000). In (Dhillon and Guan 2003), ITCC is improved to be more adaptive for high-dimensional and sparse co-occurring data.

Besides the clustering analysis, ITCC has been extended to various application scenes. It was applied to classification problems in transfer learning settings (Dai et al. 2007), which is called co-clustering-based classification (CoCC). CoCC bridges the knowledge from the source domain to that from the target domain through ITCC so that the unlabeled data in the target domain can utilize the labeled data in the source domain for classification. Furthermore, it was used to solve the classification problems in multi-view domain adaption settings using CoCC (Yang et al. 2012). ITCC was also introduced into unsupervised transfer learning (Dai et al. 2008), where the transfer clustering method utilized unlabeled data from the different distributions in the source domain to facilitate the clustering task in the target domain. With the ability to deal with sparse data, ITCC was used to settle the problems in the recommendation systems (Liang and Leng 2014). In (Bloechl, Amjad, and Geiger 2018), ITCC was developed by the combination with Markov chain aggregation.

Inspired by above work and to take full advantage of the two-sided clustering for co-occurring data in the scene of multi-view clustering, a two-sided multi-view clustering algorithm based on ITCC is proposed in the paper.

## Multi-view Information-theoretic Co-clustering

In this section, the fundamentals of ITCC are first reviewed. Then, the objective function of MV-ITCC is derived. Finally, an iterative optimization strategy of the objective function is detailed.

### Fundamentals of ITCC

ITCC is a clustering algorithm which groups the features and samples simultaneously. In general, each row of the data matrix represents a sample and each column represents a feature. Let $X$ and $Y$ be discrete random variables corresponding to the sample and feature dimensions respectively. Let $p(X,Y)$ denote the joint probability distribution between $X$ and $Y$. In practice, when $p$ is not known, it can be estimated using the normalized form of the data matrix of co-occurring data, where the random discrete variables $X$ and $Y$ can take values in the sets $\{x_1,...,x_n\}$ and $\{y_1,...,y_n\}$ respectively. Therefore, $p(x,y)$ is a definite value of the joint probability distribution.

ITCC simultaneously clusters $X$ and $Y$ respectively into $k$ and $l$ clusters. Let $k$ clusters of $X$ be $\{\hat{x}_1,\hat{x}_2,...,\hat{x}_k\}$, and $l$ clusters of $Y$ be $\{\hat{y}_1,\hat{y}_2,...,\hat{y}_l\}$. In other words, the key task of ITCC is to find the maps $\hat{X}=C_X(X)$ and $\hat{Y}=C_Y(Y)$, where $\hat{X}$ and $\hat{Y}$ are random variables that are deterministic functions of $X$ and $Y$ respectively. Let $p(\hat{X},\hat{Y})$ be the joint probability distribution between $\hat{X}$ and $\hat{Y}$, so $p(\hat{x},\hat{y})$ is a definite value of $p(\hat{X},\hat{Y})$.

Based on the two joint probability distributions before and after clustering, ITCC defines an objective function to minimize the loss of mutual information for the two distributions, that is:

$$J(\hat{X},\hat{Y}) = I(X;Y) - I(\hat{X};\hat{Y}) \quad (1)$$

where the mutual information $I(X;Y)$ quantifies the amount of information obtained about random variable $Y$ through random variable $X$, and vice versa. The purpose of (1) is to maximize the amount of information obtained about $\hat{Y}$ through $\hat{X}$ during clustering, because for a distribution $p(X,Y)$, $I(X;Y)$ is fixed. This is consistent with the information bottleneck theory which is to preserve the amount of information during data compression.

In order to simplify the solution procedure, lemma 2.1 in (Dhillon et al. 2003) has proved that (1) can be transformed to the Kullback-Leibler (KL) divergence between the two distributions before and after clustering.

$$I(X;Y) - I(\hat{X};\hat{Y}) = D(p(X,Y) \| \hat{p}(X,Y)) \quad (2)$$

where $\hat{p}(X,Y)$ can be re-expressed in the following form for convenience during optimization.

$$\hat{p}(x,y) = p(\hat{x},\hat{y})p(x|\hat{x})p(y|\hat{y})$$
$$\text{where } x \in \hat{x}, y \in \hat{y} \quad (3)$$

Through the above conversion, the objective function of ITCC is transformed to minimize the KL divergence. The detailed procedure can be found in (Dhillon et al. 2003).

## The Objective Function of MV-ITCC

In multi-view clustering, since the information of features contained in different views is usually discrepant, the clustering results for different views are inconsistent. The main task of MV-ITCC is to unify the different information from all views to achieve the optimal clustering results.

To leverage the ITCC mechanism for multi-view clustering, the most intuitive approach is to apply ITCC to the concatenated data directly. However, this would lead to two deficiencies as follows:

(1) As the features in the same view are more similar than that in different views, it is more reasonable to cluster the features within each individual view. When ITCC is applied to concatenated data, the features from different views are mixed during clustering along the feature dimension. In other words, among the values $\{\hat{y}_1,\hat{y}_2,...,\hat{y}_l\}$ of the random variable $\hat{Y}$, there may exist a $\hat{y}_i$ which is the clustering results of several features from different views. The situation is obviously not consistent with the purpose of clustering along the feature dimension.

(2) On the other hand, the simple concatenation does not reflect the different importance of individual views. Since the information contained in each view is distinct, the importance of each view for clustering is also different. A more reasonable choice is to cluster the different views individually along the feature dimension and assign each view with an adaptive weight in the objective function.

To meet the above challenges, a multi-view clustering algorithm MV-ITCC is proposed based on the mechanism of ITCC. The basic idea of MV-ITCC is as follows. Assume there are $K$ views, the data matrix of each view contains two random variables corresponding to the sample and feature dimensions respectively. Since the values along the sample dimension are consistent for all views, the discrete random variable can be denoted as $X$ for all views. On the contrary, since the values along the feature dimension are different for different views, the discrete random variables are denoted by $Y_1, Y_2, ..., Y_K$. Taking multi-language document-word co-occurrences as an example, the documents (sample dimension) for different views (different matrices) are consistent, whereas the words (feature dimension) for different views are different because the same documents can be translated into different languages.

In the procedure of clustering, the samples for all views share the same clustering results and the random variable of the clustering results along the sample dimension is denoted as $\hat{X}$. The features for the different views have their own clustering results and the random variables of individual clustering results along the feature dimension are denoted as $\hat{Y}_1, \hat{Y}_2, ..., \hat{Y}_K$. MV-ITCC exactly exploits the consistency and complementary information among views by sharing common $\hat{X}$ for all views and keeping specific $\hat{Y}_i$ for each view. To adjust the importance of different views, maximum entropy is introduced into the objective function with $w_i$ for $i$th-view. Hence, the objective function can be formalized as follows:

$$J(\hat{X},\hat{\mathbf{Y}},\mathbf{w}) = \sum_{i=1}^{K} w_i [I(X;Y_i) - I(\hat{X};\hat{Y}_i)] + \lambda \sum_{i=1}^{K} w_i \ln w_i$$
$$\text{s.t.} \quad \sum_{i=1}^{K} w_i = 1 \quad (4)$$

where $\lambda$ is a regularization parameter. The clustering results can be obtained by minimizing $J$ in (4). The first term is to minimize the weighted mutual information loss for all views and the second term is to maximize the entropy of all weights. The larger the regularization parameter $\lambda$ is, the closer the weights of different views are. Whereas the smaller the regularization parameter $\lambda$ is, the larger the weights of the important views containing effective information are. Therefore, it can balance the agreement and disagreement among views.

## Optimization of MV-ITCC

Since analytical solution to (4) is not available, an iterative strategy is adopted to optimize the objective function and the method is monotonically decreasing to minimize the objective function iteratively. First, the $K$ joint probability distributions are defined in the form of (3) for each view.

$$\hat{p}(x, y_i) = p(\hat{x}, \hat{y}_i) p(x | \hat{x}) p(y_i | \hat{y}_i)$$
$$\text{where } x \in \hat{x}, \ y_i \in \hat{y}_i, \ i = 1, 2, ..., K \quad (5)$$

According to (2), the information loss before and after clustering for each view is transformed into KL divergence. Hence, the objective function can be expressed as follows.

$$J(\hat{X}, \hat{Y}, \mathbf{w}) = \sum_{i=1}^{K} w_i D(p(X, Y_i) \| \hat{p}(X, Y_i)) + \lambda \sum_{i=1}^{K} w_i \ln w_i \quad (6)$$
$$\text{s.t.} \sum_{i=1}^{K} w_i = 1$$

$D(p(X,Y_i) \| \hat{p}(X,Y_i))$ can be represented as $Q_i$ and from the definition of KL divergence, $Q_i$ can be defined as follows.

$$Q_i = \sum_{\hat{x} \in \hat{X}} \sum_{\hat{y}_i \in \hat{Y}_i} \sum_{x \in \hat{x}} \sum_{y_i \in \hat{y}_i} p(x, y_i) \log \frac{p(x, y_i)}{\hat{p}(x, y_i)}, \ i = 1, 2, ..., K \quad (7)$$

Based on (5) - (7), the objective function of MV-ITCC can be reformed as (8).

$$J(\hat{X}, \hat{Y}, \mathbf{w}) = \sum_{i=1}^{K} w_i Q_i + \lambda \sum_{i=1}^{K} w_i \ln w_i \quad (8)$$
$$\text{s.t.} \sum_{i=1}^{K} w_i = 1$$

It can be seen from (8) that $J$ can be minimized from two aspects: by minimizing $Q_i$ for fixed $w_i$, and by minimizing $J$ with respect to $w_i$ for fixed $Q_i$. Then, the optimization of the objective function can be decomposed into two sub-problems. Hence, the problem is transformed to how to optimize $Q_i$. Next, $Q_i$ is further derived into two forms.

The joint probability distribution $\hat{p}(x, y_i)$ in (7) can be re-expressed as follows.

$$\hat{p}(x, y_i) = p(\hat{x}, \hat{y}_i) \frac{p(x)}{p(\hat{x})} \frac{p(y_i)}{p(\hat{y}_i)}$$
$$= p(y_i) \frac{p(x)}{p(\hat{x})} \frac{p(\hat{x}, \hat{y}_i)}{p(\hat{y}_i)} = p(y_i) \hat{p}(x | \hat{y}_i) \quad (9)$$
$$= p(x) \frac{p(\hat{x}, \hat{y}_i)}{p(\hat{x})} \frac{p(y_i)}{p(\hat{y}_i)} = p(x) \hat{p}(y_i | \hat{x})$$

where $\hat{p}(x, y_i)$ is expressed as two forms of conditional probability distributions. One is conditioned on the sample dimension $\hat{x}$ and another is conditioned on the feature dimension $\hat{y}_i$. $Q_i^x$ can be obtained by substituting the first form into (7).

$$Q_i^x = \sum_{\hat{x} \in \hat{X}} \sum_{\hat{y}_i \in \hat{Y}_i} \sum_{x \in \hat{x}} \sum_{y_i \in \hat{y}_i} p(x) p(y_i | x) \log \frac{p(x) p(y_i | x)}{p(x) \hat{p}(y_i | \hat{x})}$$
$$= \sum_{\hat{x} \in \hat{X}} \sum_{x \in \hat{x}} p(x) \sum_{\hat{y}_i \in \hat{Y}_i} \sum_{y_i \in \hat{y}_i} p(y_i | x) \log \frac{p(y_i | x)}{\hat{p}(y_i | \hat{x})} \quad (10)$$
$$= \sum_{\hat{x} \in \hat{X}} \sum_{x \in \hat{x}} p(x) D(p(Y_i | x) \| \hat{p}(Y_i | \hat{x})), \ i = 1, 2, ..., K$$

$Q_i^y$ is obtained by substituting the second form into (7):

---

**Algorithm 1**: MV-ITCC
**Input**: Given $K$ views, the number of clusters $C$ for $n$ samples, the convergence threshold $\varepsilon$, the number of iterations $T$ and the multi-view dataset $\chi = \{x_i\}_{i=1}^{n}$, $x_i = \{x_i^{(v)}\}_{v=1}^{K}$.
**Output**: The final clustering function $C_X(x_i)$ for sample $x_i$.
**Procedure MV-ITCC**:
1: Initialize the clustering function $C_X^{(0)}$, $C_{Y_i}^{(0)}$ for each view and initialize the weights $w_i^{(0)}$ for each view.
2: Initialize the $p(X, Y_i)$ for each view based on $\chi = \{x_i\}_{i=1}^{n}$.
3: Initialize the $\hat{p}^{(0)}(X, Y_i)$ for each view based on $p(X, Y_i)$, $C_X^{(0)}$, $C_{Y_i}^{(0)}$ and (5).
4: **For** $t \leftarrow 1, 2, ..., T$ **do**
5:   Update $C_X^{(t)}$ based on (14) with $p(X, Y_i)$, $\hat{p}^{(t-1)}(X, Y_i)$ in (10).
6:   Update $C_{Y_i}^{(t)}$ based on (15) with $p(X, Y_i)$, $\hat{p}^{(t-1)}(X, Y_i)$ in (11).
7:   Update $\hat{p}^{(t)}(X, Y_i)$ based on $p(X, Y_i)$, $C_X^{(t-1)}$, $C_{Y_i}^{(t-1)}$ and (5).
8:   Update $w_i$ for each view based on (13).
9:   Update $J^{(t)}$ with (4) and evaluate the convergence by comparing with $J^{(t-1)}$.
10: **end for**

---

$$Q_i^y = \sum_{\hat{x} \in \hat{X}} \sum_{\hat{y}_i \in \hat{Y}_i} \sum_{x \in \hat{x}} \sum_{y_i \in \hat{y}_i} p(y_i) p(x | y_i) \log \frac{p(y_i) p(x | y_i)}{p(y_i) \hat{p}(x | \hat{y}_i)}$$
$$= \sum_{\hat{y}_i \in \hat{Y}_i} \sum_{y_i \in \hat{y}_i} p(y_i) \sum_{\hat{x} \in \hat{X}} \sum_{x \in \hat{x}} p(x | y_i) \log \frac{p(x | y_i)}{\hat{p}(x | \hat{y}_i)} \quad (11)$$
$$= \sum_{\hat{y}_i \in \hat{Y}_i} \sum_{y_i \in \hat{y}_i} p(y_i) D(p(X | y_i) \| \hat{p}(X | \hat{y}_i)), \ i = 1, 2, ..., K$$

Through the above derivation, $Q_i$ can then be optimized by two steps. That is to optimize $Q_i^x$ with fixed $Y_i$ and to optimize $Q_i^y$ with fixed $X$. Therefore, integrating with the optimization of $w_i$, a three-step iterative optimization procedure for MV-ITCC is developed as follows.

First, when the clustering results along both the feature dimension and the sample dimension are fixed (i.e., $Q_i$ is fixed), Lagrangian multipliers are adopted to minimize $J$ with respect to $w_i$, $i = 1, 2, \cdots, K$. The Lagrangian function $L$ below can be derived from (8).

$$L = \sum_{i=1}^{K} w_i Q_i + \lambda \sum_{i=1}^{K} w_i \ln w_i + \gamma (\sum_{i=1}^{K} w_i - 1) \quad (12)$$

By setting $\partial L / \partial w_i = 0$, $w_i$ can be analytically expressed as (13). (*Please refer to Part 1 in the Supplementary Material for details*).

$$w_i = \frac{\exp(\frac{-Q_i}{\lambda})}{\sum_{k=1}^{K} \exp(\frac{-Q_k}{\lambda})}, \ i = 1, 2, ..., K \quad (13)$$

Second, according to (10), when the clustering results along the feature dimension are fixed (i.e., $Y_i$ is fixed), for a certain sample $x$, the optimal cluster that $x$ belongs to can be decided by the values of $Q_i^x$ when $x$ belongs to different clusters. The cluster corresponding to the smallest

$Q_i^x$ is the optimal cluster for $x$. Assume there are $K$ views for a certain $x$, the optimal cluster $C_X(x)$ for $x$ is obtained by minimizing the weighted sum of $Q_i^x$.

$$C_X(x) = \arg\min_{\hat{x} \in \hat{X}} \sum_{i=1}^{K} w_i Q_i^x \quad (14)$$

Third, in a similar way, according to (11), when the clustering results along the sample dimension are fixed (i.e., $X$ is fixed), for a certain feature $y_i$ of $Y_i$, the optimal cluster that $y_i$ belongs to can be decided by the values of $Q_i^y$ when $y_i$ belongs to different clusters. $K$ iterations are conducted in this step for the different views to obtain the optimal cluster $C_{Y_i}(y_i)$ by minimizing $Q_i^y$, $i = 1, 2, ..., K$ as follows.

$$C_{Y_i}(y_i) = \arg\min_{\hat{y}_i \in \hat{Y}_i} Q_i^y, \ i = 1, 2, ..., K \quad (15)$$

An iterative optimization algorithm is thus constructed based on (13), (14) and (15). The procedure is monotonically decreasing until the algorithm reaches convergence. The workflow of the algorithm is given in Algorithm 1.

## Experiments

### Datasets

According to (14) and (15), to optimize the objective function of the proposed method MV-ITCC*, it is necessary to know the empirical joint probability distributions of two random variables representing the sample and feature dimensions. Based on the characteristics of co-occurrences in data, the normalized form of its data matrix can be directly taken as the empirical joint probability distribution. This is because that each element of co-occurring data represents the frequency of sample-feature co-occurrence, and the co-occurrence property is consistent with the notion of joint probability.

The most common co-occurring data contains text data represented with the bag of words model (BoW) and image data represented with the bag of visual words model (BoVW) (Li and Perona 2005) or feature histograms. The BoVW model can be based on scale invariant feature transform (SIFT) (Lowe 2004), speed-up robust features (SURF), etc. The common feature histograms contain histograms of oriented gradients (HOG), local binary pattern (LBP) (Ojala et al. 1994), color histogram, etc.

Seven co-occurring datasets are used in the experiments to evaluate the effectiveness of the proposed method. Four text datasets represented with BoW model and three image datasets represented with BoVW model and feature histograms were adopted. The details of the datasets are described below and Table 1 gives the statistics of the da-

---

Table 1: Statistical of datasets

| Dataset | Size | Views/Dimensions | Clusters |
|---|---|---|---|
| Cora | 2708 | 2 (2708-1433) | 7 |
| Reuters | 1200 | 2 (2000-2000) | 6 |
| 3S | 169 | 3 (3560-3631-3068) | 6 |
| NG20 | 900 | 2 (800-800) | 3 |
| Caltech | 2033 | 2 (300-256) | 3 |
| Corel | 1000 | 2(300-256) | 10 |
| Leaves | 1599 | 2(64-64) | 100 |

tasets.

- Cora dataset: It is a dataset of publications (Zhang et al. 2014). Each publication is a sample consisting of two views: the word vector of the main content and the word vector of the reference links pointing to the publication.
- Reuters dataset: Reuters is document collection translated into five languages, where each language is regarded as a view (Jiang et al. 2012). In the experiments, English and French are selected as two views.
- 3S dataset: 3S (3Ssources) is a collection of stories gathered from three news websites (Zhang et al. 2014). Each story is a sample and the content is presented differently at the three websites. 169 stories available from all the three sources are selected as a multi-view dataset.
- NG20 dataset: NG20 is constructed from the News-Group 20 dataset according to the procedure in (Gu and Zhou 2009). 900 samples are selected randomly from three classes. Each class is divided into two sub-classes representing two views. Detailed composition of the dataset is listed in Table 2. (*Please refer to Part 2 in the Supplementary Material for details*).
- Caltech dataset: Caltech is an image dataset containing 101 classes (Kumar and Rai 2011), from which 3 classes with the largest number of samples are selected, i.e., Airplanes, Faces easy and Motorbikes. The images are preprocessed to a two-view dataset. For the first view, we extracted features using the SIFT, with which a codebook of size 300 is generated using clustering algorithm. Finally, the images are processed and represented in the form of BoVW. For the second view, LBP features are extracted and the images are represented as histogram vectors with 256 dimensions.
- Corel dataset: Corel is an image classification dataset (Jiang et al. 2012), from which 10 classes with obvious foreground objects are selected for the experiments. The same feature extraction methods used for the Caltech dataset are applied to it.
- Leaves dataset: It is an image dataset with one hundred plant species from UCI repository. There are sixteen samples for each species. Two attribute sets are selected for the experiments. They are scale margin and texture histogram.

---

* The code is available at https://github.com/DallasBuyer/MVITCC

Table 3: NMI and RI on the text datasets

| Algorithm/Dataset | Normalized Mutual Information(%) | | | | Rand Index(%) | | | |
|---|---|---|---|---|---|---|---|---|
| | Cora | Reuters | 3sources | NG20 | Cora | Reuters | 3sources | NG20 |
| MVKKM | 30.27±00 | 29.85±00 | 09.74±00 | 06.30±00 | 58.95±00 | 68.84±00 | 26.96±00 | 54.69±00 |
| WV-CoFCM | 18.34±03 | 22.44±03 | 43.95±06 | 11.38±03 | 71.33±02 | 72.79±02 | 74.51±03 | 59.47±06 |
| MVSpec | 18.56±00 | 34.77±00 | 47.44±00 | 17.05±00 | 76.58±00 | 78.64±00 | 77.56±00 | 63.64±00 |
| MVKSC | 27.67±00 | 18.51±02 | **64.94±00** | 46.51±00 | 77.09±00 | 74.43±00 | **86.15±01** | 75.25±01 |
| MV-JNMF | 29.95±02 | 26.94±00 | 40.10±04 | 20.50±02 | 74.02±01 | 76.02±00 | 73.49±02 | 58.64±01 |
| ITCC | 30.61±08 | 33.31±04 | 45.34±08 | 28.45±24 | 78.87±02 | 78.37±01 | 77.66±04 | 63.60±16 |
| K-means | 13.42±08 | 19.17±09 | 13.96±08 | 05.02±05 | 57.88±18 | 45.21±20 | 34.05±13 | 36.30±04 |
| MV-ITCC | **34.82±06** | **35.26±03** | 50.62±09 | **61.08±27** | **79.75±02** | **79.15±01** | 80.05±04 | **81.44±16** |

## Experimental Settings

To evaluate the effectiveness of the proposed method, it is compared with seven algorithms, including two single-view clustering methods and five multi-view clustering methods. The brief descriptions and parameter settings of the algorithms are as follows:

- K-means: K-means is a single-view clustering method and is regarded as the baseline for the comparison.
- ITCC: ITCC is single-view clustering method and it is used to evaluate the effectiveness of MV-ITCC. The number of clusters along the feature dimension is optimally set using the search grid $\{10,20,\cdots,100\}$.
- MVKKM: MVKKM is a K-means based multi-view clustering method inspired by unsupervised multi-kernel learning (Tzortzis and Likas 2012). The kernel width for different views is optimally set using the search grid $\{2^{-6}, 2^{-5},...,2^{0},...,2^{5},2^{6}\}$.
- WV-CoFCM: WV-CoFCM extends fuzzy K-means to the multi-view version (Jiang et al. 2015). The two regularization parameters of WV-CoFCM are optimally set using the search grid {0.01,0.1,1,10,100} and the weights of different views are optimally set using the search grid $[0,(K-1)/K]$ with the step of 0.1.
- MVSpec: MVSpec is a spectral based multi-view clustering method and integrated with multi-kernel learning (Tzortzis and Likas 2012). The setting of kernel width was the same as that in MVKKM.
- MVKSC: MVKSC is a multi-view kernel spectral clustering method formulated as a weighted kernel canonical correlation analysis (Houthuys et al. 2018). The regularization parameter is set by the search grid $\{2^{-6}, 2^{-5},...,2^{0},...,2^{5},2^{6}\}$. The setting of kernel parameter was the same as the regularization parameter.
- MV-JNMF: MV-JNMF learned a common representation matrix and different basis matrices for each view based on non-negative matrix factorization (Liu et al. 2013). In the experiments, the number of iterations is set as 30.
- MV-ITCC: For the proposed method, the regularization parameter of the maximum entropy term, the number of clusters for features and the number of iterations are all adjustable parameters. The number of clusters for the features was set in advance by fine-tuning. Based on the experiments, it was enough to set the number of iterations as 20. The regularization parameter was optimally set by using search grid $\{2^{-6}, 2^{-5},...,2^{0},...,2^{5},2^{6}\}$.

Each algorithm was executed for 30 times with different parameters to determine the best settings where the optimal performance was achieved in terms of purity, NMI (Xu, Tao, and Xu 2015) and RI (Jiang et al. 2015).

## Experimental Results

Experiments are conducted on seven multi-view datasets and the results are recorded in terms of the mean and standard deviation of purity, NMI and RI of 30 runs.

The clustering performance of text datasets in terms of NMI and RI is shown in Table 3 and that on image datasets in terms of purity, NMI and RI is shown in Table 4. It can be seen from Table 3 that the performance of MV-ITCC is considerably better than that of the other algorithms. Although the performance of ITCC is inferior to that of MV-ITCC, it still outperforms the other algorithms in the whole since ITCC is suitable for the high-dimensional and sparse text data. The performance on the image datasets in Table 4 also shows the effectiveness of MV-ITCC. But the advantages on the image datasets are not so obvious than that on the textual datasets. From the table 3 and Table 4, it can be concluded that 1) the performance of the algorithms relies heavily on the datasets and 2) it benefits to conduct two-sided clustering for multi-view co-occurring data.

On the other hand, it can be seen clearly from Table 3 that the performance of K-means, MVKKM and WV-CoFCM is poor, due to the fact that MVKKM and WV-CoFCM are multi-view algorithms extended from K-means. They need dense sampling and are not suitable for high-dimensional and sparse data, especially for the 3S and NG20 with limited samples. On the contrary, the performance of these three K-means based algorithms is preferable on the image datasets. This is because that these three image datasets satisfy the dense sampling with low dimension of features and large number of samples.

Table 4: Purity, NMI and RI on the image datasets

| Algorithm/Dataset | Purity(%) | | | Normalized Mutual Information(%) | | | Rand Index(%) | | |
|---|---|---|---|---|---|---|---|---|---|
| | Corel | Caltech | Leaves | Corel | Caltech | Leaves | Corel | Caltech | Leaves |
| MVKKM | 33.60±00 | 79.98±00 | **81.43±00** | 25.95±00 | 48.86±00 | **92.03±00** | 80.81±00 | 76.25±00 | 99.24±00 |
| WV-CoFCM | 38.86±01 | 77.77±03 | 60.66±01 | 26.10±01 | 55.30±03 | 80.56±00 | 84.15±02 | 78.63±03 | 99.06±00 |
| MVSpec | 40.90±00 | 89.18±00 | 68.29±00 | 27.94±00 | 65.82±00 | 82.66±00 | 84.62±00 | 86.68±00 | 99.10±00 |
| MVKSC | 31.71±00 | 59.81±00 | 23.91±01 | 19.24±00 | 26.88±00 | 52.02±01 | 82.69±00 | 64.42±00 | 97.98±00 |
| MV-JNMF | 41.08±02 | 95.57±00 | 73.73±02 | 28.37±01 | 85.39±00 | 87.57±01 | 84.80±00 | 94.76±00 | 99.19±00 |
| ITCC | 40.78±02 | 97.63±00 | 77.46±02 | 29.59±02 | 90.54±01 | 91.44±01 | 84.64±00 | 97.08±00 | 99.32±00 |
| K-means | 38.09±02 | 84.35±00 | 76.70±02 | 27.18±01 | 56.07±00 | 90.68±01 | 83.83±01 | 80.43±00 | 99.29±00 |
| MV-ITCC | **41.96±02** | **98.39±00** | 78.79±02 | **30.14±01** | **92.09±00** | 92.00±01 | **84.86±00** | 97.99±00 | **99.36±00** |

For the dataset NG20, where the views have low correlation, the performance of MV-ITCC is considerably better than that of all the other algorithms, even for ITCC. This indicates that it is not appropriate to apply clustering methods directly to the concatenated data with low correlation among views, while the proposed MV-ITCC can effectively handle this situation. When applying the clustering algorithm directly to the concatenated data, different views are treated equally. However, when there is low correlation among views, the information contained in each view has a big difference and the proposed multi-view clustering algorithm can balance the importance of different views.

To further verify the effectiveness of MV-ITCC, statistical analysis is conducted on all multi-view algorithms by Friedman test followed by post-hoc test. (*Please refer to Part 3 in the Supplementary Material for details*). The results indicate that the proposed method is significantly superior to the other multi-view clustering algorithms except for the MV-JNM F. Meanwhile, it can be seen from Table 3 and Table 4 that the performance of MV-ITCC is better than that of MV-JNMF to some extend although the improvement is not statistically significant.

## Parameter Study

There are three adjustable parameters for MV-ITCC. The first one is the number of iterations of the iterative strategy in Algorithm 1. The second one is the regularization parameter of maximum entropy term in (8). The third one is the number of clusters during the clustering process along the features dimension. They are analyzed as follows.

Refer to Figure 1, the regularization parameter is analyzed by investigating the variation in the weights and the corresponding NMI values. The *x*-axis represents the sequence number of the regularization parameters. Figure 1 clearly demonstrates the effect of the regularization parameter. It can be seen that when the regularization parameter $\lambda \to 0$, one of the two views plays a more significant role while the effect of another view can almost be neglected. Here the disagreement among views is excessively excavated and the more important view dominates the cluster-ing results. When $\lambda \to \infty$, the importance of two views is almost equal. Here the agreement is overemphasized and the complementary information among views is not leveraged sufficiently. When $\lambda$ is set as the seventh, sixth and fourth parameters for the Corel, Caltech and Leaves datasets respectively, the NMI reaches the optimal results where the agreement and disagreement gets a balance.

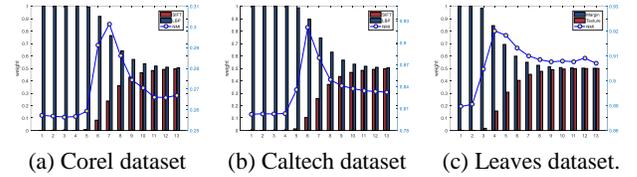

(a) Corel dataset    (b) Caltech dataset    (c) Leaves dataset.

Figure 1: Regularization parameter.

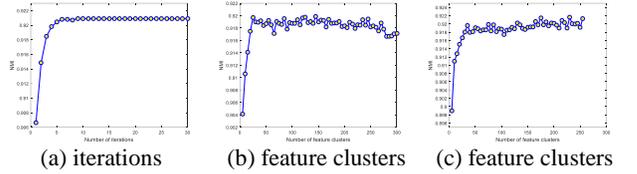

(a) iterations    (b) feature clusters    (c) feature clusters

Figure 2: Parameter analysis on the Caltech dataset.

Refer to Figure 2(a), the NMI values is plotted along with the variation of the number of iterations on the Caltech dataset. It can be seen that MV-ITCC converges very well after 8 iterations and NMI performance reaches the highest which indicates the property of good convergence.

The number of feature clusters for each view is a flexible parameter. Figure 2(b) illustrates NMI values under different number of feature clusters of the first view with that of the second view fixed. The fixed feature clusters of the second view are set as the number obtained by fine-tuning on the second view. Similarly, Figure 2(c) illustrates NMI values under different number of feature clusters of the second view with that of the first view fixed and the fixed clusters number is obtained by the same way. NMI performance is stable and the fluctuation range of NMI is less than 0.005 when the number of feature clusters is set greater than 50. So MV-ITCC is insensitive to the parameter of the number of feature clusters to some extent.

## Conclusion

The main contribution of the paper is that a novel two-sided multi-view clustering algorithm based on ITCC is proposed. It is inspired by taking the advantage of co-clustering in the scene of multi-view clustering for the co-occurrences in data. The choice of multi-view clustering algorithms is heavily dependent on the application scenes. For the scenes involving co-occurrence data, two-sided clustering can be effective. Most of existing multi-view clustering methods only focus one-sided clustering along the sample dimension and do not take advantage of the characteristics of the datasets. MV-ITCC make up for the shortcomings of the traditional multi-view clustering algorithms for this type of data. Besides the co-occurring data, it is a significant research to introduce different clustering techniques into the multi-view clustering for the other types of data.

## Acknowledgements

This work was supported in part by the National Key Research Program of China (2016YFB0800803), the NSFC (61772239), the Jiangsu Province Outstanding Youth Fund (BK20140001), the National First-Class Discipline Program of Light Industry Technology and Engineering (LITE2018-02), and Basic Research Program of Jiangnan University Key Project in Social Sciences JUSRP1810ZD.